\documentclass[10pt,twocolumn,letterpaper]{article}

\usepackage{cvpr}              % To produce the CAMERA-READY version

\usepackage{multirow}
\usepackage{bbding}
\usepackage{newtxmath}
\usepackage{algorithm}
\usepackage{algorithmic}
\usepackage[dvipsnames]{xcolor}

\usepackage[symbol]{footmisc}

\newcommand{\tablestyle}[2]{\setlength{\tabcolsep}{#1}\renewcommand{\arraystretch}{#2}\centering\footnotesize}
\newlength\savewidth

\usepackage{times}
\usepackage{epsfig}
\usepackage{graphicx}
\usepackage{amsmath}
\usepackage{amssymb}
\usepackage{booktabs}
\usepackage{multirow}
\usepackage{tabularx}
\usepackage{enumitem}
\usepackage{gensymb}
\usepackage{colortbl}
\usepackage{caption}
\usepackage{makecell}
\usepackage{wrapfig}
\usepackage{amssymb}% http://ctan.org/pkg/amssymb
\usepackage{pifont}% http://ctan.org/pkg/pifont

\definecolor{mycyan}{cmyk}{.1,0,0,0}
\newcommand{\cmark}{\ding{51}}%
\newcommand{\xmark}{\ding{55}}%

\newcommand\mypara[1]{\vspace{0mm}\noindent\textbf{#1}}
\definecolor{cvprblue}{rgb}{0.21,0.49,0.74}
\usepackage[pagebackref,breaklinks,colorlinks,citecolor=cvprblue]{hyperref}

\def\paperID{6861}
\def\confName{CVPR}
\def\confYear{2024}

\title{MLP Can Be A Good Transformer Learner}
\makeatletter
\newcommand{\printfnsymbol}[1]{%
  \textsuperscript{\@fnsymbol{#1}}%
}
\makeatother
\author{
Sihao Lin$^1$\footnotemark[1] \quad Pumeng Lyu$^2$\footnotemark[1]  \quad Dongrui Liu$^{2,3}$ \quad Tao Tang$^4$ \quad
Xiaodan Liang$^{4,5,7}$ \\
Andy Song$^1$\quad Xiaojun Chang$^{6,7}$\footnotemark[2] \\ 
{\normalsize $^1$RMIT University  
\; $^2$Shanghai AI Laboratory
\; $^3$Shanghai Jiao Tong University
\; $^4$Shenzhen Campus of Sun Yat-sen University
} \\
{\normalsize 
\quad $^5$DarkMatter AI Research
\quad $^6$University of Technology Sydney \quad $^7$MBZUAI}\\
{\tt \small \{linsihao6,trent.tangtao,xdliang328\}@gmail.com \quad \{lvpumeng,liudongrui\}@pjlab.org.cn} \\ 
{\tt \small andy.song@rmit.edu.au \quad xiaojun.chang@uts.edu.au }
\vspace{-6mm}
}
\begin{document}
% \maketitle

\twocolumn[{ 
\maketitle
  \centering
  \includegraphics[width=1.0\textwidth]{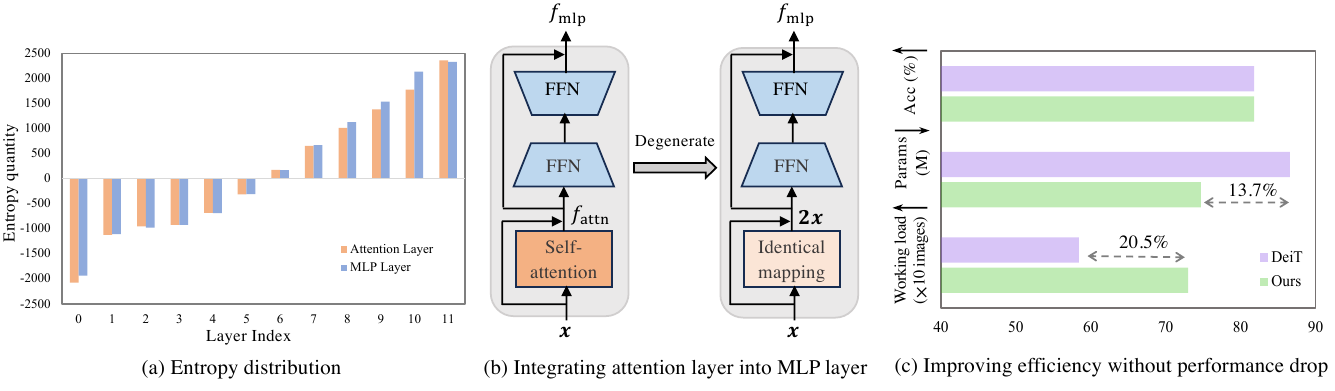}
  \vspace{-6mm}
  \captionof{figure}{ 
  \textbf{Pruning the attention layer from the perspective of entropy.}
  (a) 
  We use entropy to illustrate the information amount carried out by the attention layers and MLP layers (\ie two FFN layers) in each transformer block of DeiT-B~\cite{touvron2021training}. 
  We observe that the entropy quantity of the bottom blocks is lower than that of the top blocks. 
  We identify a pattern that, the attention layer with low entropy is accompanied by the MLP layers with the entropy quantity at the same level.
  (b) In the bottom blocks, MLP layers can elicit the information as much as that of the attention layers. On the other hand, they are under-exploited given the low entropy quantity compared to those MLP layers in the top blocks. We thus propose to integrate the uninformative attention layer into its subsequent MLP layer through proper optimization.
  (c) As a result, our method can reduce 13.7\% parameters of DeiT-B and improve 20.5\% working load in the same memory budget without performance degradation.
  }
\label{fig:fig1}
\vspace{3mm}
}]

\renewcommand{\thefootnote}{\fnsymbol{footnote}}
\footnotetext[1]{Equal contribution.}
\renewcommand{\thefootnote}{\arabic{footnote}}
\renewcommand{\thefootnote}{\fnsymbol{footnote}}
\footnotetext[2]{Corresponding author.}
\renewcommand{\thefootnote}{\arabic{footnote}}
%%%% 2nd ver %%%%

\begin{abstract}
Self-attention mechanism is the key of the Transformer but often criticized for its computation demands. Previous token pruning works motivate their methods from the view of computation redundancy but still need to load the full network and require same memory costs. This paper introduces a novel strategy that simplifies vision transformers and reduces computational load through the selective removal of non-essential attention layers, guided by entropy considerations. We identify that regarding the attention layer in bottom blocks, their subsequent MLP layers, i.e. two feed-forward layers, can elicit the same entropy quantity. Meanwhile, the accompanied MLPs are under-exploited since they exhibit smaller feature entropy compared to those MLPs in the top blocks. Therefore, we propose to integrate the uninformative attention layers into their subsequent counterparts by degenerating them into identical mapping, yielding only MLP in certain transformer blocks. Experimental results on ImageNet-1k show that the proposed method can remove 40\% attention layer of DeiT-B, improving throughput and memory bound without performance compromise\footnote{\url{https://github.com/sihaoevery/lambda_vit}}.
\end{abstract}
\vspace{-6mm}

\section{Introduction}
\vspace{-1mm}
\label{sec:intro}

Vision Transformer~\cite{dosovitskiy2020image,touvron2021training} is becoming dominant for vision tasks~\cite{weng2023mask,han2023html,han2023shot2story20k}. It is believed that self-attention mechanism is the key component for its success, which models the dense similarity between two entries. Nonetheless, researchers have found that the attention layer is redundant~\cite{michel2019sixteen,bian2021attention}, \eg attention maps across different heads~\cite{liu2023efficientvit} or stages~\cite{bhojanapalli2021leveraging} might be similar to each other. 
To this end, a broad array of works~\cite{dutson2023eventful,pan2021ia,rao2021dynamicvit,tang2022patch,bolya2022token,chen2021psvit,wei2023joint,liang2022not,xu2022evo} 
propose to prune/merge the redundant tokens to reduce computation redundancy. Nonetheless, these methods still need to load the full network and consume the same memory costs as the original model. 

To this end, this work aims to directly remove those uninformative attention layers to push the memory bound.
We investigate this problem from the perspective of entropy, which measures the information quantity of a network. 
As a motivator, we visualize the entropy distribution of the attention layers, together with their subsequent MLP layers, of DeiT-B~\cite{touvron2021training} as illustrated in~\cref{fig:fig1} (a). 
Specifically, one can observe that in the bottom blocks, the entropy quantity of the attention layer is 
lower than that of the top blocks.
In particular, we identify a pattern that, the attention layer with low entropy is accompanied by the MLP layers with the entropy quantity at the same level. Our finding brings a novel perspective to the inefficient attention layers. On one hand, since MLP layers in bottom blocks contain the entropy as same as the attention layers, they may elicit the same information. On the other hand, these MLPs are under-exploited and thus can be optimized to
be as expressive as those MLPs in the top blocks. Therefore, 
a natural question is raised:
Can we \textit{integrate the uninformative attention layer into its subsequent MLPs}? 

More concretely, in the context of entropy, we question whether the information carried out by the attention layer can be transplanted into the corresponding MLP layer, through proper optimization.
As shown in~\cref{fig:fig1} (b),
the output feature of the attention layer is the input of the subsequent MLP layer. Given this fact, we propose a simple dilution learning technique that gradually degenerates the attention layer into identical mapping.
Eventually, the resulting identical mapping together with the residual connection can be integrated into the subsequent MLPs, yielding only MLPs in certain Transformer blocks. 

\begin{figure}[!t]
  \vspace{0mm}
  \centering
    \includegraphics[width=1.0\linewidth]{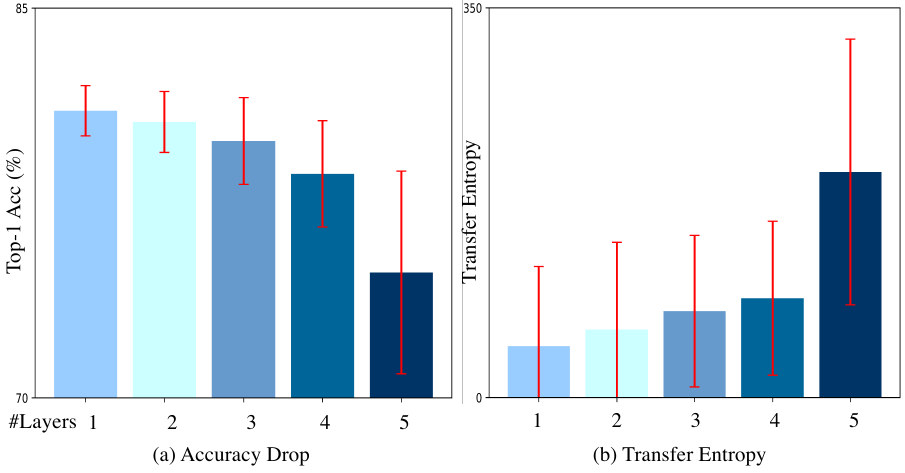}
  \caption{
  \textbf{Interaction of multiple layers.}
  Both figures have the same $x$-axis (\#Layer). We use the idea of transfer entropy to measure the interaction on multiple layers. Here, we randomly mask 1$\sim$5 attention layers of a pre-trained DeiT-B. We record the means (bars) and variances (red lines) of model performances in (a) and the corresponding transfer entropies in (b). It is clear that model performance drops as transfer entropy increases in both mean and variance. As a motivator, we aim to remove attention layers with fewer interactions (\ie transfer entropy).
  }
  \label{fig:interaction}
  \vspace{-4mm}
\end{figure}

Another question is which attention layers should be selected for consequent manipulation. 
Probably it is natural to perform on the consecutive bottom blocks since they carry out less information. However, such a strategy neglects the potential interaction among different layers. 
For instance, we randomly mask $N$ attention layers ($N$ is from 1 to 5) of a pre-trained DeiT-B and repeat this process over 20 times. We therefore get the means (bars) and variances (red lines) of model performances (~\cref{fig:interaction} (a)) and the corresponding transfer entropies (~\cref{fig:interaction} (b)) when removing 1$\sim$5 layers. Model performance drops as transfer entropy increases in both mean and variance, indicating the importance of interaction among multiple layers.

To this end, we propose the E\textbf{n}tr\textbf{o}py-based \textbf{S}election Strat\textbf{e}gy, dubbed as \textbf{NOSE}, to identify the combination of different attention layers that cause minimum impact on the consequent performance. Specifically, we use transfer entropy to approximate the interaction between an ordered array of attention layers and the final output layer. ~\cref{fig:interaction} (b) shows that the transfer entropy has a great variation across different combinations. 

We validate the proposed method on three benchmarks ImageNet-1k~\cite{deng2009imagenet}, CIFAR-100~\cite{krizhevsky2009learning}, and ADE20k~\cite{Zhou_2017_CVPR}. 
The experimental result evidences that our framework can effectively discard uninformative attention layers and learn the robust feature without performance compromise. For instance, our method removes 40\% attention layers of DeiT-B without performance drop in ImageNet-1k.
To summarize, we claim the following contribution in this work:
\begin{itemize}
    \item We propose a novel framework that transplants the knowledge of non-essential attention layers into their subsequent MLP layers.
    \item From the perspective of transfer entropy, we propose the Entropy-based Selection Strategy to identify the correlation between an ordered array of attention layers and the final output layer, which causes less or even no degradation to the network performance.
    \item We propose a simple yet effective dilution learning technique that degenerates attention layers into identical mapping layers. Eventually, the identical mapping together with the residual connection are taken as the input of the MLP layer, yielding only MLP in certain blocks.
\end{itemize}
\section{Related Work}
It is acknowledged that the concept of Transformer is proposed by Vaswani \etal~\cite{vaswani2017attention} for natural language process. Lately, Dosovitskiy \etal~\cite{dosovitskiy2020image} introduce the vision transformer for image recognition. Since its emergence, the \textit{self-attention} efficiency regarding the quadratic complexity has engaged considerable interest from industrial and research community. Existing methods motivate this problem from two perspectives: token aggregation and token pruning.

\mypara{Token aggregation.}
There are works that approximate the full attention with partial attention by leveraging the locality. SwimT~\cite{liu2021swin} and FocalT~\cite{yang2021focal} use local windows to extract the feature from neighbor tokens, resulting in the feature map of smaller resolution. MetaFormer~\cite{yu2022metaformer} and PSViT~\cite{chen2021psvit} apply simple pooling operation among local tokens to reduce the length of the token array. Recently, ToMe~\cite{bolya2022token} proposes to gradually combine two similar tokens by bipartite soft matching without needing to train.

\mypara{Token pruning.} 
Some work aims to dynamically prune the uninformative token during training. DynamicViT~\cite{rao2021dynamicvit} uses a prediction module to measure the importance score for each token and progressively prunes the redundant tokens stage by stage. Rather, Patch Slimming~\cite{tang2022patch} performs token pruning in a top-down manner. It identifies the valuable tokens in the last layer and in turn requires the previous layer to discriminate these tokens from the redundant one. EViT~\cite{liang2022not} simply identifies the attentive tokens given the similarity with the classification token. Then the inattentive tokens are fused into a supplement token. Evo-ViT~\cite{xu2022evo} proposes the Fast-slow Token Evolution where valuable tokens and uninformative tokens are separately updated using different strategies. Recently, TPS~\cite{wei2023joint} proposes the Join Pruning and Squeezing module that first identifies the reserved tokens and pruned tokens, which are fused into the reserved tokens according to their matching score. 

Yet, existing token pruning methods discussed above require the same memory cost as the original model since they are compelled to the full network architecture and even additional modules. Our method can push the limit of memory bound since we combine attention layers with subsequent MLP layers and remove self-attention architectures.

\section{Methods}
\vspace{-2mm}
We first briefly introduce the preliminaries of vision transformer in~\cref{sec:preliminary}. We use entropy to quantify the information carried out by the attention layer (~\cref{sec:entropy}) and propose the selection strategy to identify which layers are supposed to be removed (\cref{sec:nose}). In~\cref{sec:sparse}, we present a simple network \textit{dilution} recipe that gradually degenerates the attention layer into identical mapping.

\subsection{Preliminary}
\label{sec:preliminary}
Vision transformer (ViT) is first introduced by Dosovitskiy \etal~\cite{dosovitskiy2020image} for image classification~\cite{deng2009imagenet}. A ViT is composed of a patch embedding layer $\mathcal{P}$ and a stack of transformer blocks $\mathcal{A}$, following a task-specific head $\mathcal{G}$. 
\begin{equation}
\begin{aligned}
{\rm ViT}&=\mathcal{G} \circ \mathcal{A} \circ \mathcal{P}, \\
\mathcal{A} &= A_{l} \circ \cdots \circ A_{2} \circ A_{1}.
\end{aligned}
\end{equation}
Given a predefined patch size $h \times w$, the patch embedding layer encodes an image $I \in \mathbb{R}^{H\times W \times 3}$ into $P=H/h\times W/w$ patch tokens with dimension $d$. It is then prepended the classification token to form the image tokens, which are fed into the transformer blocks. Typically, a transformer block includes a self-attention layer Attn and a subsequent MLP layer (\ie two feed-forward layers).  
Consider a transformer block in $\mathcal{A}$:

\begin{equation}
\begin{aligned}
f_{\rm attn}&= {\rm Attn}(\boldsymbol{x})+\boldsymbol{x}, \\
{\rm Attn}(\boldsymbol{x}) &= {\rm softmax}(\frac{Q\cdot K^{\top}}{\sqrt{d}})\cdot V, \\
    Q=W_{Q}(\boldsymbol{x}),\ &K=W_{K}(\boldsymbol{x}),\ V=W_{V}(\boldsymbol{x}).
\label{eq:attn_output}
\end{aligned}
\end{equation}
Here $\boldsymbol{x}=\{x_i\}\in \mathbb{R}^d$ is the input tokens for classification token $i=0$ and patch tokens $1\leq i \leq P$. $W_{Q}$, $W_{K}$ and $W_{V}$ are the linear projections that projects $\boldsymbol{x}$ to query $Q$, key $K$ and value $V$ of size $(P+1)\times d$. By convention, a residual connection is applied to the output of Attn, and the Layer Norm (LN) result~\cite{ba2016layer} is fed into MLP, generating the output of this block:
\begin{equation}
\begin{aligned}
    % x_{i}^{\rm mlp} = {\rm MLP}({\rm LN}(x_{i}^{\rm attn}))+x_{i}^{\rm attn}.
    f_{\rm mlp} = {\rm MLP}({\rm LN}(f_{\rm attn}))+f_{\rm attn}.
\end{aligned}
\label{eq:output}
\end{equation}

\subsection{Entropy Quantification}
\label{sec:entropy}
By definition, entropy~\cite{guan2019towards} can be used to measure the information quantity of a network. 
Accordingly, one can calculate the entropy of a certain layer given the probability of its feature:
\begin{equation}
H(F) = -\int_{}^{} p(f) {\rm log}p(f) \,df, f\in F.
\label{eq:shannon}
\end{equation}
Nonetheless, it is difficult to directly measure the probability distribution of a feature map: $p(f),f\in F$. 
Following~\cite{sun2022entropy, Siam2020}, we use the Gaussian distribution as the probability distribution of the intermediate feature in a layer. Therefore, the entropy of a certain layer is approximated as the mathematical expectation of $F \sim \mathcal{N}(\mu,\,\sigma^{2})$:
\begin{equation}
    \begin{split}
     H(F) & = -\mathbb{E}[{\rm log}\mathcal{N}(\mu, \sigma^{2})] \\
     & = -\mathbb{E}[{\rm log}[(2 \pi \sigma^2)^{-1/2} {\rm exp}(-\frac{1}{2\sigma^2}(f - \mu)^2)]] \\
     & = {\rm log}(\sigma) + \frac{1}{2}{\rm log}(2\pi) + \frac{1}{2},
    \end{split}
\label{eq:gaussian}
\end{equation}
\noindent where $\sigma$ is the standard deviation of the feature set $f\in F$. Typically, a batch of images is passed into a vision transformer to obtain the feature set $F$ of the attention layer and MLP layer (\cref{eq:attn_output} \& (\ref{eq:output})), respectively. $H(F)$ is proportional to ${\rm log}(\sigma)$ plus two additional constants. Without loss of generality, the two constants are neglected in the following analysis. In practice, we apply~\cref{eq:gaussian} to each channel of the intermediate feature. Then, without considering constant terms, the entropy of each layer is proportional to the summation of logarithm of standard deviation of each feature channel:

\begin{equation}
    H(F) \propto H_{\sigma}(F) = \sum_{j}^{} {\rm log}[\phi(F^j)].
\label{eq:entropy}
\end{equation}

\noindent Thus, $H_{\sigma}(F)$ is the value proportional to the entropy of a layer, either attention or MLP layer. $\phi(F^{j})$ calculates the standard deviation of $j^{th}$ channel of the feature set $F$.

\subsection{Interaction among Multiple Attention Layers}
\label{sec:nose} 
The above discussion formulates the entropy of a single layer. 
Our goal is to remove an ordered array of attention layers that are less significant to the original architecture.
As shown in~\cref{fig:fig1} (a), it is plausible to remove the attention layers in the bottom blocks with relatively low entropy. However, such a strategy largely neglects the potential interaction across different layers, which is proved to be important in~\cref{fig:interaction}.

As a remedy, we resort to the \textit{transfer entropy} (TE)~\cite{zhang2023transferable,schreiber2000measuring,nb2022causality} that measures the information amount of directed transfer between two layers.
Given a target layer, transfer entropy compares the difference in entropy quantity in the presence and absence of the source layer. 
\begin{equation}
    TE = H(F_{\rm target}) - H(F_{\rm target}|\mathcal{A}\backslash \{{\rm Attn}_{\rm source}\}).
\label{eq:trans_entropy}
\end{equation}
Here $H(F_{\rm target})$ is the original entropy of the target layer defined in~\cref{eq:entropy}. We compute the entropy $H(F_{\rm target}|\mathcal{A}\backslash \{{\rm Attn}_{\rm source}\})$ in the condition that source attention layer ${\rm Attn}_{\rm source}$ is masked out, \ie set to identical mapping. Hence, the numeric value of $TE$ can reflect the significance of the source layer over the target layer, measuring their correlation. 
We aim to identify the combination of multiple attention layers that have the minimum correlation with the final output layer of the network.

Therefore, we propose the E\textbf{n}tr\textbf{o}py-based \textbf{S}election Strat\textbf{e}gy, dubbed as NOSE, to select the attention layers with minimum transfer entropy to the final output layer.
The proposed NOSE will measure the transfer entropy between the attention layers and the final output layer iteratively.
At each round, NOSE traverses the candidate attention layers $\mathcal{C}$ and figures out the layer has a minimum transfer entropy using greedy search. This layer is appended to the state set $\mathcal{S}$, which will be detached from the candidate set and won’t participate in the next loop. We then repeat the procedure by taking into account the previous state till the combination reaches a sufficient amount.

\subsection{Integrating Attention Layer into MLP}
\label{sec:sparse}
Given the fact that MLP layer would take as input the output of the attention layer, our method degenerates the attention layers into identical mapping. Hence, the identical mapping and the associated residual connection, can be integrated into the subsequent MLP layer, yielding only MLP in the transformer block.

\mypara{Diluting the attention output.}
Following~\cite{guo2020learning,louizos2018learning}, an attention layer is decoupled to the original architecture and a sparse mask.
The ~\cref{eq:attn_output} is reformulated as:
\begin{equation}
    f_{\rm attn}= M \odot {\rm Attn}(\boldsymbol{x})+\boldsymbol{x}, \ M\in \mathbb{R}^{(P+1)\times d} ,
\label{eq:naive_decay}
\vspace{-5mm}
\end{equation}
where $\odot$ is element-wise multiplication. The sparse mask $M$ is usually subject to some constraints, \eg L$_0$ norm~\cite{louizos2018learning,guo2020learning}, and is used to regularize the sparsity of the attention output. In our case, $M$ is initialized as 1
and is manually decayed till 0
along the training process. We showcase in the experiments that the implementation of $M$ is robust to different choices. Once the sparse mask is decayed to 0, the output $f_{\rm attn}$ of the attention layer becomes the residual connection. 

\begin{algorithm}[t]
    \renewcommand{\algorithmicrequire}{\textbf{Input}}
	\renewcommand{\algorithmicensure}{\textbf{Output}}
    \caption{Training Procedure of Our Method}
    \label{alg:ours}
    \begin{algorithmic}[1]
        \REQUIRE  
        a ViT,
        state set $\mathcal{S}$, candidate set $\mathcal{C}$, amount of selecting layers $N$, training set [$\mathcal{I}$,$\mathcal{Y}$], decay function $D$, sparse mask $M$, training iterations $T$, loss function $\mathcal{L}$.
        \ENSURE simplified ViT with $N$ attention layer get removed.
        \STATE $\mathcal{S}\leftarrow \emptyset$, $\mathcal{C}\leftarrow \{{\rm Attn}_1,{\rm Attn}_2,...,{\rm Attn}_l\}$, $M \leftarrow 1$
        \vspace{1mm}
        \STATE Identify the combination of attention layers:
        \FOR { $n=0,1,2,...,N$}
        \STATE Traverse the attention layer in candidate set $\mathcal{C}$: \\
        $\underset{i}{\arg\min}(TE(\mathcal{S}\cup\{ {\rm Attn}_i \}, \mathcal{G}))$ 
        \; \; \;\;\,\,\,$\triangleleft$ greedy search
        \STATE $\mathcal{S}\leftarrow \mathcal{S}\cup\{ {\rm Attn}_i \}$,
        $\mathcal{C}\leftarrow \mathcal{C}\backslash \{{\rm Attn}_i\}$
        $\triangleleft$ update state
        \ENDFOR
        \vspace{1mm}
        \STATE Diluting the attention layers:
        \FOR {$t=0,1,2,...,T$}
        \STATE  Fit a batch of data [$I$,$Y$] sampled from [$\mathcal{I}$,$\mathcal{Y}$]:\\
        minimize $\mathcal{L}({\rm ViT}(I),Y)$ 
        $\triangleleft$ apply~\cref{eq:feat_com} on $\mathcal{S}$
        \STATE $M \leftarrow D(M)$ 
        \quad \quad \quad \quad \, $\triangleleft$ decay sparse mask
        \ENDFOR
    \end{algorithmic}
\end{algorithm}

\mypara{Feature compensation.}
As the sparse mask is decayed, it continuously vanishes the gradient of the attention layer. Hence, the backward gradient of the degenerated output will be smaller than the original one, which incurs training instability. 
To this end, we propose the feature compensation, which adaptively compensates the gradient loss brought by sparse mask:
\begin{equation}
\begin{aligned}
    f_{\rm attn}&= M \odot {\rm Attn}(\boldsymbol{x})+(1-M)\odot \boldsymbol{x}+\boldsymbol{x}\\
    &= M \odot {\rm Attn}(\boldsymbol{x})+(2-M)\odot \boldsymbol{x}.
\end{aligned}
\label{eq:feat_com}
\vspace{-2mm}
\end{equation}

Here, we introduce a new term $(1-M)\odot \boldsymbol{x}$ compared to~\cref{eq:naive_decay}. It will correspondingly compensate the loss of attention output ${\rm Attn}(\boldsymbol{x})$ following the pace of $M$. Eventually, the attention layer is degenerated to an identical mapping, resulting in the output $2\boldsymbol{x}$. As a result, the attention layer is integrated into the subsequent MLP layer and is no longer required in the inference stage. We summarize our pipeline in~\cref{alg:ours}. 
\vspace{-1mm}
\section{Experiment}
\vspace{-1mm} 

\begin{figure}[!t]
  \vspace{0mm}
  \centering
    \includegraphics[width=1.0\linewidth]{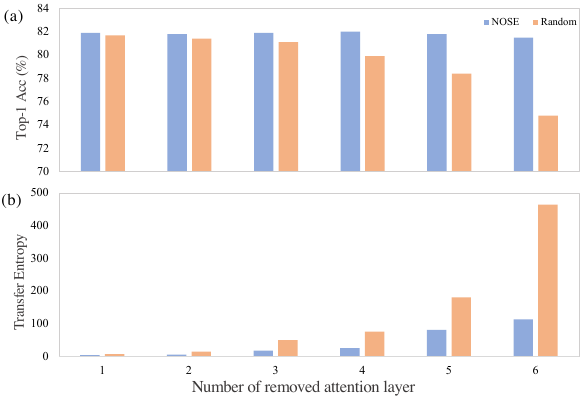}
\vspace{-5mm}
  \caption{
  \textbf{NOSE \vs Random selection.}
  (a) NOSE consistently outperforms the random selection on ImageNet-1k. (b) This is because NOSE can identify the attention layers with less interaction with the final output layer, which is reflected by transfer entropy.}
  \label{fig:vsrandom}
\vspace{-5mm}
\end{figure}

\subsection{Baseline Setting}
\label{sec:exp_setup}
\vspace{-2mm}
\noindent \textbf{Benchmark.} CIFAR-100~\cite{krizhevsky2009learning} is an image classification benchmark with 100 semantic categories. The training set and validation set have 50,000 and 10,000 samples, respectively. ImageNet-1k~\cite{deng2009imagenet} is a challenging classification dataset with 1,000 categories. It has more than 1\textbf{M} training samples and 50,000 validation samples. ADE20K~\cite{Zhou_2017_CVPR} is a semantic segmentation dataset with 150 classes, which has 20,000 training samples and 2,000 validation samples. 

\mypara{Evaluation protocol.}
We first assess our method on ImageNet-1k. 
Furthermore, we verify the proposed method on ADE20k for dense classification. To evaluate the feature richness learned by our method, we perform the transfer learning on CIFAR-100 using the weights pre-trained on ImageNet-1k. 

\mypara{Implementation Details.} Given the popularity and influence, we adopt the DeiT~\cite{touvron2021training} as the implementation of the ViT, which has the standard softmax-attention layer. Throughout the paper, we perform experiments on the network architecture DeiT-B. We adopt the training recipe of DeiT on ImageNet-1k and CIFAR-100. For ImageNet-1k, we decay the sparse mask $M$ from 1 to 0 for 300 epochs. We follow the experimental setting of TinyMIM~\cite{ren2023tinymim} on ADE20k. We provide more details and additional experiments on other backbones in~\cref{sec:more_exp} of the appendix.

\subsection{Main Result}
\label{sec:main_result}
\noindent \textbf{Validating the entropy-guided selection strategy.} 
We inspect the effectiveness of the proposed selection strategy NOSE.
We compare the method against the random selection scheme where the same amount of attention layers are sampled randomly. 
Due to the high space complexity,
we sample three times from the feasible combinations $\mathcal{C}_{N}^{n}$.  

The result on ImageNet-1k demonstrated the effectiveness of the proposed NOSE. 
As illustrated in~\cref{fig:vsrandom} (b), our method proved to identify the combination of attention layers with lower transfer entropy compared to random selection. As a result, the proposed NOSE would cause less or even no degradation to the performance.
Specifically, in ~\cref{fig:vsrandom} (a), when a few attention layers (\eg 1 or 2) are removed, the random selection scheme would not affect the vision transformer too much. 
On the other hand, when increasing the amount of removed attention layers, the network would suffer from the random selection scheme since it is likely to select the inappropriate combination, declining the classification result. For instance, when randomly selecting 4 out of 12 attention layers, the network performance would deteriorate from 81.8\% to 79.9\%. 
In contrast, the proposed NOSE can properly identify the attention layers with less transfer entropy and consequently preserve the performance. 
We can observe that the proposed NOSE is able to remove 5 out of 12 attention layers of DeiT-B without performance compromise. Additionally, when half of the attention layers are removed, our method slightly declines the performance by 0.3\% while random selection would lead to a drastic drop of 7\%, demonstrating the effectiveness of the proposed NOSE.
We also implement the First-$N$ baseline which removes the first $N$ consecutive attention layers in~\cref{tab:first_n} of the appendix.

\begin{table*}[t]
    \centering
\caption{Compare to other methods on ImageNet-1k. We report the performance, throughput, and memory bound. $^{*}$ means using training. $^{\dag}$ means a more aggressive configuration.}
\vspace{-3mm}
\resizebox{2.0\columnwidth}{!}
{\tablestyle{7pt}{1.0}
    \begin{tabular}{@{}lcccccc@{}}
    \toprule
         Method &Top-1 (\%) &Top-5 (\%)&FLOPs (G) &Throughput (images/s)&Params (M) &Memory bound (images/10GB)\\ \hline
         Deit-B~\cite{touvron2021training} &81.8 &95.6 &17.6 &299 &86.6  &606\\
         \hline
         DynamicViT~\cite{rao2021dynamicvit} &81.3 &- &11.5 &464 &89.5  &606 \\
         Evo-ViT~\cite{xu2022evo} &81.3 &- &11.7 &474 &87.3  &608 \\
         EViT~\cite{liang2022not} &81.3 &95.3 &11.5 &458 &86.6 &608 \\
         TPS~\cite{wei2023joint} &81.4 &- &11.5 &468 &89.5  &606 \\
         ToMe~\cite{bolya2022token} &80.6 &- &11.5 &462 &86.6 &606\\
         ToMe$^{*}$~\cite{bolya2022token} &81.4 &- &11.5 &462 &86.6 &606\\
         DiffRate~\cite{chen2023diffrate} &81.5 &- &11.5 &465 &86.6 &606\\
         \hline
         Ours(40\%) &\textbf{81.8} &\textbf{95.6} &15.0 &390 &74.7 ($\downarrow$13.7\%) &730 ($\uparrow$20.5\%)\\
         Ours(40\%)+ToMe &81.6 &95.4&11.4  &478 &74.7 ($\downarrow$13.7\%) &730 ($\uparrow$20.5\%) \\
         Ours(40\%)+ToMe$^{\dag}$ &81.4 &95.3 &\textbf{10.9} &\textbf{507} &74.7 ($\downarrow$13.7\%) &730 ($\uparrow$20.5\%) \\ \hline
         Ours(50\%) &81.5 &\textbf{95.6} &14.5 &408 &\textbf{72.4} ($\downarrow$16.4\%) &\textbf{732}($\uparrow$20.8\%) \\
         Ours(50\%)+ToMe &81.3 &95.4 &11.9 &462 &\textbf{72.4} ($\downarrow$16.4\%) &\textbf{732}($\uparrow$20.8\%) \\
    \bottomrule
    \end{tabular}}
    \label{tab:imagenet_comp}
\vspace{-4mm}
\end{table*}
\begin{table}[t]
    \centering
\caption{Transfer learning on CIFAR-100.}
\vspace{-3mm}
\resizebox{0.8\columnwidth}{!}
{\tablestyle{15pt}{1.0}
    \begin{tabular}{@{}lcc@{}}
    \toprule
         {Method} &Fine-tuning &Linear probing \\ \hline
         Deit-B~\cite{touvron2021training}      &\textbf{90.5} & 80.6 \\
         \hline
         Evo-ViT~\cite{xu2022evo}     &90.1 &79.1 \\
         EViT~\cite{liang2022not}     &90.0 &80.2 \\
         TPS~\cite{wei2023joint}      &90.1 &76.5 \\
         Ours(40\%)  & \textbf{90.3} & \textbf{ 81.3} \\
         Ours(50\%)  & 90.2  &80.6\\
    \bottomrule
    \end{tabular}
    }
    \label{tab:cifar}
\vspace{-5mm}
\end{table}

\mypara{Comparison on ImageNet-1k.} We compare our method with other works on ImageNet-1k. As illustrated on~\cref{tab:imagenet_comp}, our method showcases competitive performance compared to token pruning method. For instance, our method exceeds TPS by 0.4\% and EViT by 0.5\% regarding Top-1 Acc. 

The issue of memory bound remains untouched in current token pruning methods ~\cite{xu2022evo,wei2023joint}, yet it is important for compact devices with limited memory budget. Without bells and whistles, we record the maximum amount of input images during inference till the model fills up 10GB budget of the GPU memory. 
Since our method unloads the attention layer, it has a considerable reduction in model size. For instance, removing 40\% attention layers can lead to a reduction of 13.7\% regarding the network parameters, as shown in ~\cref{tab:imagenet_comp}. Consequently, our model consumes less memory and eliminates the issue of memory bound, improving more than 20\% working load compared to other methods. We test the throughput in a V100 (32GB) GPU with batch size 128. When removing 50\% attention layers, our method improves the throughput by a margin of 36.5\% (408 \vs 299). In addition, our method can combine with the unsupervised token merging method (\eg~\cite{bolya2022token}) seamlessly and further improve the throughput by 69.6\% (507 \vs 299) while maintaining a competitive performance. Given these results, our method can boost both the throughput and memory bound, bringing the best of two worlds.

\mypara{Transfer learning on CIFAR-100.}
We assess the transferable ability of the learned feature from ImageNet-1k to CIFAR-100. The experiments are conducted in two protocols:
1) \textit{Fine-tuning:} The backbone is initialized with the pre-trained weights from ImageNet-1k and updated through end-to-end training. 2) \textit{Linear probing:} The learned feature from ImageNet-1k is frozen and only a linear classifier (\ie a full-connected layer plus a softmax layer) is trained.

As illustrated in~\cref{tab:cifar}, for the setting of fine-tuning, our method slightly outperforms other comparison methods and is close to the original DeiT-B. In particular, when it comes to linear probing, the proposed method can exceed other methods by a clear margin. For instance, when removing 50\% attention layer, our model surpasses TPS~\cite{wei2023joint} and Evo-ViT~\cite{liang2022not} by 4.1\% and 1.5\%, respectively. This is because token pruning methods implicitly encode the dataset bias in order to discriminate the useful tokens from the redundant ones. Thus, their learned representations exhibit less generalization ability to unseen datasets. 

\begin{table}[t]
    \centering
\caption{Results on ADE20k. $^{*}$ means using 2$\times$ training iterations.}
\vspace{-3mm}
\resizebox{1.0\columnwidth}{!}
{\tablestyle{15pt}{1.1}
    \begin{tabular}{@{}lccc@{}}
    \toprule
         Method & mIoU (\%) &mAcc (\%) &aAcc (\%)\\ \hline \hline
         \multicolumn{4}{c}{\textit{From scratch}} \\ \hline
         Deit-B~\cite{touvron2021training} &24.4 &32.3 &71.0 \\
         EViT~\cite{liang2022not} &24.0 &32.2 &70.5\\
         TPS~\cite{wei2023joint} &23.5 &31.7 &70.5\\
         Our (40\%) &\textbf{24.6} &\textbf{32.5} &\textbf{71.0} \\
         Ours (50\%) &23.9 &31.9 &70.6 \\ \hline
         Deit-B$^*$~\cite{touvron2021training} &26.2 &33.4 &72.3\\
         EViT$^*$~\cite{liang2022not} &25.7 &\textbf{33.5} &71.9 \\
         TPS$^*$~\cite{wei2023joint} &25.1 & 33.1 &71.6 \\
         Ours$^*$ (40\%) &\textbf{26.1} &33.4 &\textbf{72.3} \\
         Ours$^*$ (50\%) &25.6 &33.3 &71.9 \\ \hline \hline
         \multicolumn{4}{c}{\textit{Pre-trained on ImageNet-1k}} \\
         \hline
         Deit-B~\cite{touvron2021training} &47.0 &57.5 &82.6 \\
         EViT~\cite{liang2022not} &45.5 &55.9 &81.9 \\
         TPS~\cite{wei2023joint} &45.3 &55.1 &81.9\\
         Ours (40\%) &\textbf{46.2} & \textbf{56.5} & \textbf{82.2}\\
         Ours (50\%) &45.6 & 55.2 &82.0\\ \hline
         Deit-B$^*$~\cite{touvron2021training} &48.2 &58.4 &83.1 \\
         EViT$^*$~\cite{liang2022not} &46.7 &57.1 &82.4 \\
         TPS$^*$~\cite{wei2023joint} &46.4 &56.9&82.1\\
         Ours$^*$ (40\%) &\textbf{47.5} &\textbf{57.7} &\textbf{82.7} \\
         Ours$^*$ (50\%) &46.7 &57.3 &82.2 \\
    \bottomrule
    \end{tabular}
    }
    \label{tab:ade20k}
\vspace{-6mm}
\end{table}

\mypara{Result on ADE20k.}
We generalize the proposed framework to the task of dense prediction at ADE20k, which is rarely explored by previous work~\cite{xu2022evo,liang2022not,wei2023joint}. 
By default, the model is trained for 160k iterations, and the sparse mask $M$ is decayed at every single iteration. 
As illustrated in~\cref{tab:ade20k}, when training from scratch, our method reduces 40\% attention layer while maintaining the performance, exhibiting its application in real-world scenarios. In addition, our method consistently outperforms other comparison methods~\cite{liang2022not,wei2023joint} in terms of the mIoU metric. A possible reason is that token pruning method explicitly drops the uninformative tokens and thus loses the global context, which is crucial for dense classification tasks. Even though TPS and EViT would re-utilize these pruned tokens, they might undermine the global dependency between tokens.

When the model is pre-trained on ImageNet-1k. All the model's performance is greatly boosted. In this case, our model with 40\% attention layer removed shows a gap ($\sim$ 0.7\%) compared to the baseline. We conjecture that baseline model can learn a better global dependency from ImageNet-1k. Again, our method consistently outperforms other methods.

\begin{figure}[!t]
  \vspace{0mm}
  \centering
    \includegraphics[width=1.0\linewidth]{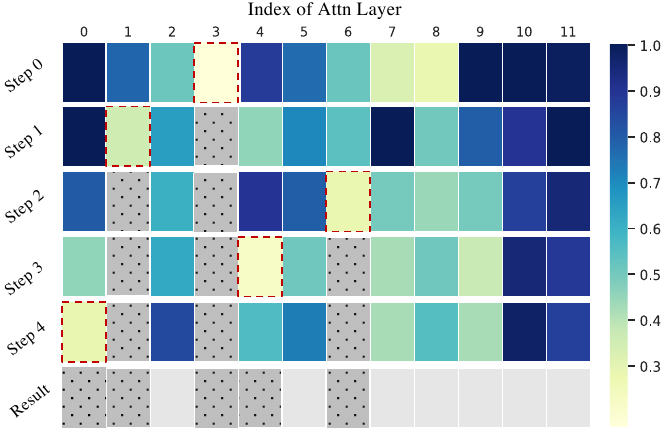}
  \caption{
  \textbf{Visualization of the proposed NOSE.} 
  For each step, a row visualizes the transfer entropy, normed to [0,1], of each attention layer associated with the final output layer. We use greedy search to select the one with minimum transfer entropy, denoted by the {\color{red}red} dashed box, \eg, layer 3 is selected at step 0. The selected layer is denoted by a {\color{gray} gray} dotted box and is suspended to a state set. In the next step, NOSE repeats this procedure on the rest attention layers considering the previous state. Finally, the attention layer indexed by [0,1,3,4,6] will be integrated into their subsequent MLP layers.
  }
  \label{fig:nose}
  \vspace{-4mm}
\end{figure}

\mypara{Visualization of NOSE.} 
We visualize the trajectory of the proposed NOSE where it identifies 5 out of 12 attention layers for elimination in~\cref{fig:nose} on ImageNet-1k. Each row represents the transfer entropy of the attention layers (\ie source layer) related to the final output layer of the vision transformer (\ie target layer). The greedy search is applied to select the attention layer, denoted by the red dashes box, with minimum transfer entropy at each step. In the next step, the selected layer of the previous step is denoted by a gray dotted box and suspended to the state set. NOSE repeats to calculate the transfer entropy for each candidate layer by taking into account the previously selected layers. 

It is counterfactual that, though layer 0 has the least entropy (\cref{fig:fig1} (a)), NOSE would select layer 3 at the first step. This is because NOSE would consider the interaction between layers rather than treat them separately. In the
early stage, NOSE would select the layers that are not consecutive to each other. We conjecture that two consecutive attention layers would result in a complex interaction towards the final output layer. Thus, in the beginning, NOSE tends to select the interval layers. As for the layers at the top blocks, they also have a complex interaction with the output layer since they often learn the high-level semantics that are significant to the output layer. Finally, the attention layers indexed by [0,1,3,4,6] are identified as the combination that has the least interaction with the output layer.

\subsection{Ablation Study and Sensitivity Analysis} 
\noindent \textbf{Sensitivity of the sparse mask $M$.}
In~\cref{eq:feat_com}, we introduce the sparse mask $M$ that is used to dilute the attention layer. For ImageNet-1k, we adopt the linear decay from 1 to 0 with 300 epochs in the main experiments. For ADE20k, it is decayed linearly at each iteration. Here we investigate the robustness of $M$ on ImageNet-1k using different step sizes and implementations, where a quantity of 40\% attention layers are selected to be integrated into the MLPs. As shown in~\cref{tab:lambda}, when the decay epoch is set to 300, both cosine and linear function would result in the same performance, implying the robustness of our methods. Not surprisingly, reducing the decay epochs, \ie increasing the decay step size, will provoke a slight performance drop. In contrast, decreasing the step size will stabilize the training process and lead to a minor improvement. 

\begin{table}[]
    \centering
    \caption{Ablation study on sparse mask.}
\resizebox{1.0\columnwidth}{!}
{\tablestyle{10pt}{1.0}
    \begin{tabular}{@{}cccc@{}}
    \toprule
         Function &  Decay Epoch &Top-1 Acc. (\%) &Top-5 Acc. (\%)\\ 
         \hline
         \multirow{3}{*}{Linear} 
         & 200 &81.4 &95.6 \\
         & 300 &81.8 &95.6 \\
         & 400 &82.0 &95.7 \\ \hline
         \multirow{3}{*}{Cosine} 
         & 200 &81.5 &95.5 \\
         & 300 &81.8 &95.6 \\
         & 400 &81.9 &95.6 \\ 
    \bottomrule
    \end{tabular}}
    \label{tab:lambda}
\vspace{-0mm}
\end{table}
\begin{table}[t]
    \centering
    \caption{Ablation on feature compensation.}
    \vspace{-2mm}
\resizebox{1.0\columnwidth}{!}
{\tablestyle{5pt}{1.2}
    \begin{tabular}{@{}cccc@{}}
    \toprule
         Remove ratio&Feat. compensation& Top-1 Acc. (\%) &Top-5 Acc. (\%) \\ \hline
         \multirow{2}{*}{40\%} & \xmark &81.4 &95.6 \\
         &\cmark &81.8 &95.6 \\ \hline
         \multirow{2}{*}{50\%} &\xmark &80.9 &95.4 \\
         &\cmark &81.5 &95.6 \\
    \bottomrule
    \end{tabular}
    }
    \label{tab:feat_comp}
\vspace{-1mm}
\end{table}

\mypara{Ablating the feature compensation.}
~\cref{eq:naive_decay} naively diminishes the attention layer by applying the sparse mask. In the end, only the residual connection will be forwarded to the subsequent MLP layers. However, since the backward gradient of the attention layer becomes smaller as long as the sparse mask is decayed, it would incur instability for training. As a remedy, we introduce the feature compensation in~\cref{eq:feat_com}. Finally, the attention layer is degenerated to an identical mapping. We conduct the ablation study to validate the effectiveness of the proposed feature compensation. The result is shown in~\cref{tab:feat_comp}. We observe that feature compensation can consistently improve the consequent performance on ImageNet-1k. The more attention layers are removed, the more benefit it can bring.

\begin{figure*}[!t]
  \vspace{0mm}
  \centering
    \includegraphics[width=1.0\linewidth]{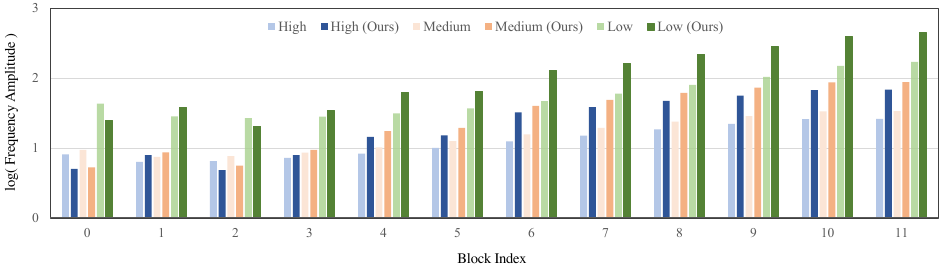}
    \vspace{-6mm}
  \caption{ 
  \textbf{Visualization of feature frequency.} We analyze feature expressivity from the frequency perspective. We apply Discrete Fourier Transform to the output feature of each block, where frequency domain is divided into low, medium, and high components. From blocks 3 to 11, our model encodes more significant high-frequency components compared to DeiT-B, implying superior feature power~\cite{bai2022improving,guo2023aloft}.
  }
  \label{fig:fft}
  \vspace{-2mm}
\end{figure*}

\begin{figure}[!t]
  \vspace{0mm}
  \centering
    \includegraphics[width=1.0\linewidth]{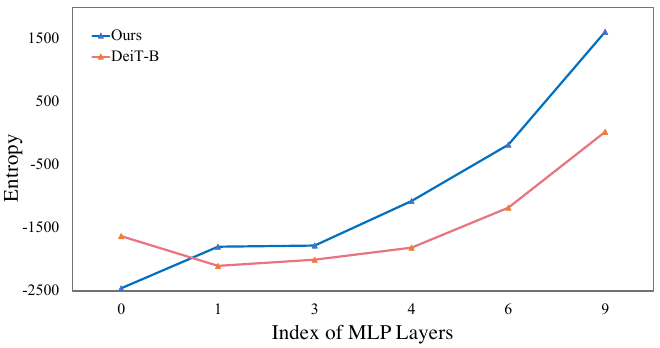}
  \caption{
  \textbf{Entropy of MLP layer at the pruned index.} Compared to the original DeiT-B, the MLP layers of our method can lead to a high entropy quantity at the pruned index [0,1,3,4,6,9].
  }
  \label{fig:after_entropy}
\vspace{-0mm}
\end{figure}

\mypara{Comparison of entropy of the MLP layers.} Our work proposes that the attention layers with low entropy quantity can be integrated into their subsequent MLP layers. Consequently, the MLP layers are expected to be more expressive in order to compensate for the reduction of attention layers. We investigate this property by comparing the entropy quantity of MLP layers at the pruned index (~\cref{sec:main_result}). Specifically, 
our model removes the 6 out of 12 attention layers indexed by [0,1,3,4,6,9]. We measure the entropy of the corresponding MLP layers in ImageNet-1k. The result in~\cref{fig:after_entropy} shows that the MLP layers of our method can surpass the original DeiT-B in terms of the entropy metric by a large margin, evidencing that they are more informative.

\mypara{Removal rates.} 
We investigate the removal rates on DeiT-B in~\cref{tab:ratio} of the appendix.

\section{A Look at Feature Expressivity}
In~\cref{tab:imagenet_comp} and ~\cref{tab:cifar}, our method and the comparison methods exhibit comparable performance on ImageNet-1k as well as the CIFAR-100 in the setting of fine-tuning. Nonetheless, when it turns to linear probing on CIFAR-100, the comparison methods lag behind our method by a substantial margin. Though removing 40\% attention layers, our method can even surpass the original DeiT-B by 0.7\% (81.3\% \vs 80.6\%).
Given this finding, we are interested in analyzing the feature space learned by our method. 

To this end, we aim to analyze the representation power of the DNN from a frequency perspective \cite{rahaman2019spectral, xu2018understanding, lin2023deep, guo2023aloft}. Specifically, we apply the Discrete Fourier Transform (DFT) to the output feature of each transformer block on ImageNet-1k. The frequency domain is divided into low [0,0.3$\pi$), medium [0.3$\pi$,0.7$\pi$) and high [0.7$\pi$,$\pi$] components. ~\cref{fig:fft} shows that the DNN trained by the proposed method encodes more significant high-frequency components in top blocks, \ie, the high-frequency component's strength of the proposed method is greater than that of the DeiT-B from block 3 to 11. Although previous studies indicate that high-frequency components are more difficult and slower to be encoded by DNNs \cite{rahaman2019spectral, xu2018understanding, liu2023towards, zhou2023concept}, the proposed method enforces the DNN to encode more high-frequency components. Furthermore, high-frequency components are useful for generalization ability \cite{guo2023aloft, bai2022improving}. Combining these previous findings and experimental observations, we may explain the effectiveness of the proposed method. \emph{I.e.}, reducing 40\% parameters and encoding more significant high-frequency components does not lead to performance degradation.
\section{Conclusion}
This work aims to remove the attention layers from the perspective of entropy. In particular, we propose the entropy-guided selection strategy (NOSE) to measure the interaction among multiple layers, which identifies the combination of attention layers that has the least influence on the model outputs. Then, we gradually degenerate those attention layers into identical mapping using a dilution learning technique, yielding only MLP in those transformer blocks.
We demonstrate the effectiveness of our method on ImageNet-1k, ADE20k, and CIFAR-100 by comparing it to current state-of-the-art strategies.
Our method reduces the network parameters as well as memory requirements. Therefore, it is able to increase the working load, which remains untouched by previous token pruning methods. Combined with the unsupervised token merging method, it strikingly boosts the throughput of the vision transformer. We also discuss the learned features of our model through DFT. 
The result shows that compared to the original DeiT-B, our model's feature map has a significant amplitude in the high-frequency components, implying superior feature power.
\section{Acknowledgement}
\vspace{-2mm}
This work was supported in part by the National Science and Technology Major Project under Grant No. 2020AAA0109704, Guangdong Outstanding Youth Fund (Grant No. 2021B1515020061), and Australian Research Council (ARC) Discovery Program under Grant No. DP240100181. The authors thank Yuetian Weng (@MonashU) for discussions.

\clearpage
{
    \small
    \bibliographystyle{ieeenat_fullname}
    \bibliography{main}
}

\clearpage
\setcounter{page}{1}
\maketitlesupplementary

\section{Code Asset}
\mypara{Acknowledgement.}
In~\cref{sec:exp_setup}, we introduce the used benchmarks. The code of this work is built upon previous works (~\cref{tab:code_asset}). The authors thank their open sourcing.

\section{Performance May Not Show The Full Picture.}
In~\cref{sec:nose}, we adopt the idea of transfer entropy and propose the NOSE to measure the interaction between an ordered array of attention layers and the final output layer.
The associated combination of attention layers with minimum transfer entropy is selected for removal. 

\begin{figure}[t!]
  \vspace{0mm}
  \centering
    \includegraphics[width=1.0\linewidth]{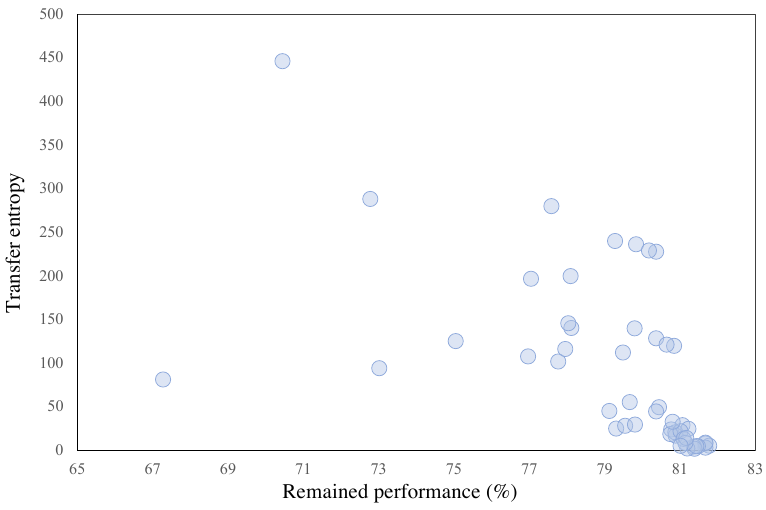}
  \caption{
  Correlation between remained performance and transfer entropy. Each point is a combination of attention layers with two metrics: transfer entropy and remained performance.
  }
  \label{fig:performanc_te}
  \vspace{-4mm}
\end{figure}

One can mask certain attention layers (\ie set to identical mapping) and measure the performance, namely \textit{remained performance}. This metric is plausible to reflect the interaction between the corresponding attention layers and the final output layer, where higher remained performance indicates less interaction. We argue that the remained performance does not show the full picture of the network. We sample some combinations of attention layers and visualize their transfer entropy together with the remained performance in~\cref{fig:performanc_te}. We find that two metrics are in part correlated. Specifically, most of the points are scattered on the right side. Typically, a combination with lower transfer entropy has a higher remained performance. In contrast, given several combinations with the same remained performance, their transfer entropy varies largely. Since transfer entropy is more consistent, we use it to determine the correlation among multiple layers. We also perform a case study to show the superiority of transfer entropy against the remained performance in~\cref{tab:case_study}. Although layer index [0,1,3,4,6] has a lower remained performance compared to layer index [1,2,3,4,6], the resulting performance is more favorable.

\begin{figure}[t!]
  \vspace{0mm}
  \centering
    \includegraphics[width=1.0\linewidth]{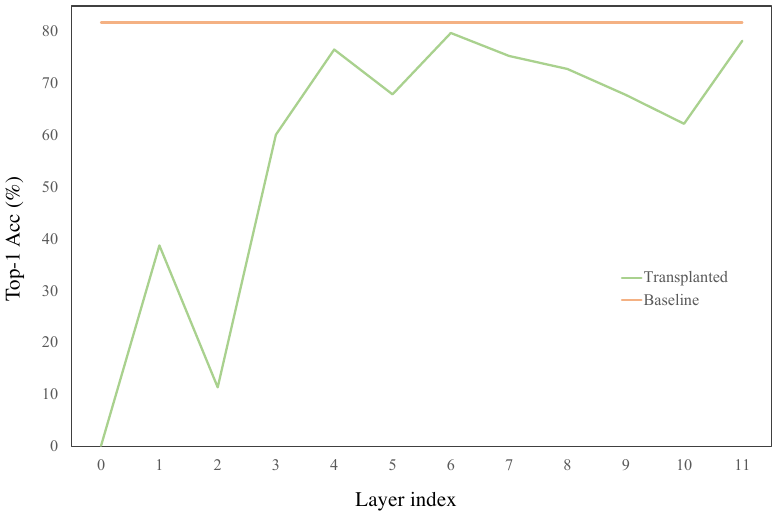}
  \caption{
  We transplant the transformer blocks of the original DeiT-B into the corresponding blocks of our model and measure the performance to investigate feature compatibility. The top blocks of our model are more compatible with the original model.
  }
  \label{fig:transplant}
  \vspace{-7mm}
\end{figure}

\begin{table*}[!t]
    \centering
    \caption{Used code asset in our work.}
    \resizebox{2.0\columnwidth}{!}
{\tablestyle{7pt}{1.1}
    \begin{tabular}{l|l|c|c}

    \toprule
         Exp.& \multicolumn{1}{c|}{URL} &Version & License  \\ \hline
         \multirow{2}{*}{ImageNet-1k} & \url{https://github.com/facebookresearch/deit} &263a3f & Apache-2.0  \\
         &\url{https://github.com/huggingface/pytorch-image-models} &0.3.2 &Apache-2.0 \\
         CIFAR-100&\url{https://github.com/facebookresearch/ToMe} &af95e4 &Creative Commons  \\ \hline
         ADE20k &\url{https://github.com/OliverRensu/TinyMIM} &d08470 &NA \\ \hline
         FLOPs &\url{https://github.com/facebookresearch/fvcore} &9d683a &Apache-2.0 \\
    \bottomrule
    \end{tabular}
    }
    \label{tab:code_asset}
\end{table*}
\begin{table*}[h]
    \centering
    \caption{Case study regarding transfer entropy and remained performance. }
    \begin{tabular}{c|cccc}
    \toprule
         Removed index & Transfer entropy $\downarrow$ & Remained performance (\%)$\uparrow$ &Top-1 (\%)$\uparrow$ &Top-5 (\%)$\uparrow$ \\ \hline
         $[1,2,3,4,6]$ &446.0 &\textbf{70.45} &81.2 &95.4 \\
         $[0,1,3,4,6]$ &\textbf{81.2}  &67.28 &\textbf{81.8} &\textbf{95.6} \\
    \bottomrule
    \end{tabular}
    \label{tab:case_study}
\end{table*}

\begin{table*}[!t]
    \centering
    \caption{More experiments on DeiT-S and DeiT-T. The number in brackets indicates the ratio of attention layers removed.}
\resizebox{2.0\columnwidth}{!}
{\tablestyle{7pt}{1.05}
    \begin{tabular}{l|ccccc}
    \toprule
         Method& Top-1 (\%)$\uparrow$ &FLOPs (G)$\downarrow$&Params (M)$\downarrow$ &Throughput (images/s)$\uparrow$ &Memory bound (images/10GB)$\uparrow$ \\ \hline \hline
         DeiT-S~\cite{touvron2021training} (baseline)  &79.9 &4.6 &22.1 &1318 &1168 \\
         Evo-ViT~\cite{xu2022evo} &79.4 &3.0 & 22.1 &1914 &1168 \\
         EViT~\cite{liang2022not} &79.5 &3.0 &22.1 &1921 &1168 \\
         ToMe~\cite{bolya2022token}     &79.5 &2.9 &22.1 &1905 &1168 \\
         DiffRate~\cite{chen2023diffrate} &79.6 &2.9 &22.1 &1805 &1168 \\
         TPS~\cite{wei2023joint} &79.7 &3.0 &22.1 &1896 &1168 \\ \hline
         Ours (25\%) &\textbf{80.1} &4.2 &\textbf{20.3} &1502 &\textbf{1382} \\ 
         Ours (30\%) &\textbf{79.8} &4.0 &\textbf{19.7} &1588 &\textbf{1388} \\ 
         Ours (40\%) &79.6 &3.9 &\textbf{19.1} &1648 &\textbf{1392} \\ \hline
         Ours (25\%)+ToMe &\textbf{79.9} &\textbf{2.7} &\textbf{20.3} &\textbf{2128} &\textbf{1352}\\
         Ours (30\%)+ToMe &79.6 &3.0 &\textbf{19.7} &\textbf{1932} &\textbf{1354}\\
         \hline \hline
         DeiT-T~\cite{touvron2021training} (baseline) &72.2 &1.3 &5.7 &3487 &2320\\
         % DynamicViT~\cite{rao2021dynamicvit} &71.4 &0.8 &5.9 &5279 & 2600 \\
         EViT~\cite{liang2022not} &71.9 &\textbf{0.8} &5.7 &5178 &2320 \\
         Evo-ViT~\cite{xu2022evo} &72.0 &\textbf{0.8} &5.9 &5258 &2320 \\
         ToMe~\cite{bolya2022token} &71.2 &0.9 &5.7 &4508 &2320 \\
         ToMe~\cite{bolya2022token} &70.9 &\textbf{0.8} &5.7 &4949 &2320 \\
         TPS~\cite{wei2023joint} &\textbf{72.3} &\textbf{0.8} &5.7 &5012 &2320 \\ \hline
         Ours (25\%)&\textbf{72.5} &1.1 &\textbf{5.3} &4001 &\textbf{2610} \\
         Ours (30\%)&71.9 &1.1 &\textbf{5.1} &4196 &\textbf{2610} \\ \hline
         Ours (25\%)+ToMe&\textbf{72.3} &\textbf{0.8} &\textbf{5.3} &\textbf{5313} &\textbf{2600} \\
         Ours (30\%)+ToMe&71.7 &0.9 &\textbf{5.1} &4846 &\textbf{2604} \\ 
    \bottomrule
    \end{tabular}
    }
    \label{tab:small}
\vspace{-3mm}
\end{table*}

\section{More Experiments}
\label{sec:more_exp}
\mypara{More details.} 
For ImageNet-1k, we use 8 GPUs with a batch size of 128 per GPU. The learning rate is set to 1e-3 and a cosine scheduler is used to regulate the learning rate till it reaches at 1e-5. We use the AdamW optimizer where beta=(0.9,0.999). For CIFAR-100, we adpot a batch size of 384 for each GPU. The image resolution is resized to 224$\times$224. And we use the SGD optimizer with a learning rate 0.1. For ADE20k, we also use 8 GPU and each GPU processes 2 input images. The optimizer is SGD and learning rate is 0.01. The polynomial scheduler with power 1.0 is used to decay the learning rate at each iteration.

\vspace{3mm}
\mypara{Feature space compatibility.} 
We are interested in the feature space learned by our method. Inspired by network transplant~\cite{yang2022deep}, we propose to \textit{transplant} the original transformer blocks, indexed from 0 to 11, of a pre-trained Deit-B into the corresponding blocks of our model. 
The classification accuracy is used to measure the compatibility.
As shown in~\cref{fig:transplant}, we find that our model, starting from block 3, is more compatible with the feature space learned by the full architecture in the top blocks. We conjecture that in bottom blocks indexed by [0,1,2], transformer would learn low-level semantics that are not very generalized. In particular, even when the attention layers are removed, blocks 4 and 6 exhibit high compatibility, indicating our model learns the feature space close to the original architecture.

\begin{figure*}[!t]
  \vspace{0mm}
  \centering
    \includegraphics[width=1.0\linewidth]{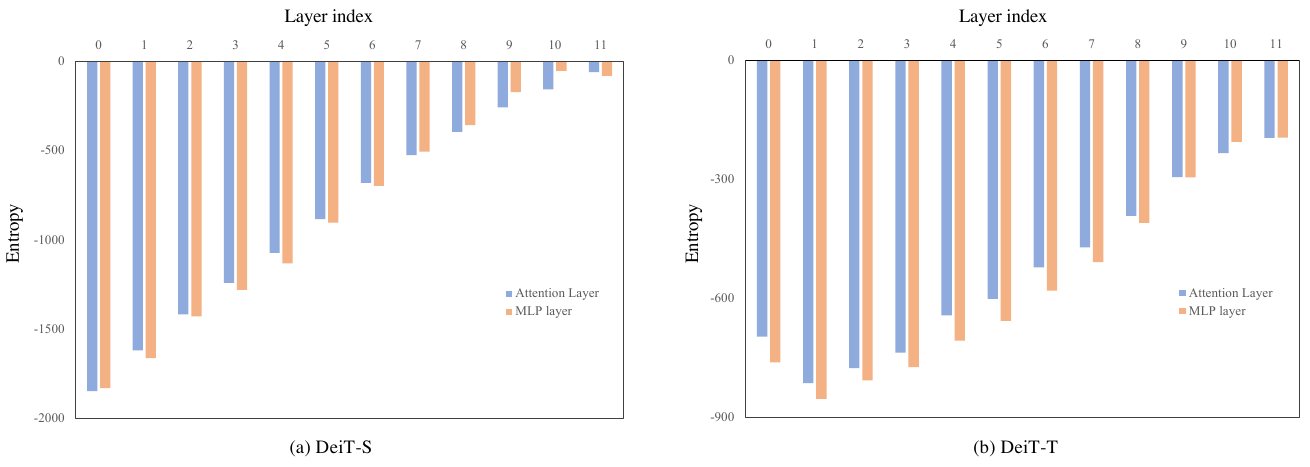}
    \vspace{0mm}
  \caption{
  Entropy distribution of DeiT-S and Deit-T. We observe that the two distributions have a similar pattern to that of DeiT-B. 
  }
  \label{fig:entropy_st}
  \vspace{-0mm}
\end{figure*}

\vspace{3mm}
\mypara{More backbones.} We assess our model on two additional backbones: DeiT-S and DeiT-T. We visualize their entropy distribution in~\cref{fig:entropy_st}. We observe that the two entropy distributions have a similar pattern to that of DeiT-B. 
The number in the brackets indicates the ratio of attention layers removed. As shown in~\cref{tab:small}, for DeiT-S, our method generally improves the memory bound by $\sim$18.5\% and the throughput\footnote{Measured on a RTX 3090 GPU with batch size 256.} by 19.5\% . When cooperated with an unsupervised token merging method, our method, while removing 25\% attention layers, can further improve the throughput by 54\% and outperforms other methods without performance compromise. Note that when combined with token merging, the working load of our model slightly decreases. This is because the tensor manipulation introduced by token matching will consume a quantity of memory~\cite{liu2023efficientvit}.
A similar experiment result is observed for DeiT-T.
\vspace{-2mm}

\begin{table}[!h]
    \centering
    \caption{Removing first $N$ attention layers on DeiT-B.}
    \vspace{-2mm}
\resizebox{1.0\columnwidth}{!}
{\tablestyle{6pt}{1.0}
    \begin{tabular}{@{}cl|cccccccc@{}}
    \toprule
           \multicolumn{2}{c|}{Remove Num.}&1 &2& 3& 4 &5 &6 &7\\ \hline
         \multirow{2}{*}{First-$N$}
         &T.E. & 140 & 167 & 211 & 333 & 498 & 636 & 645 \\
         &Top-1 (\%) &81.8 & 81.8 & 81.7 & 81.4 & 80.8 & 79.8 & 77.6 \\ \hline
        \multirow{2}{*}{NOSE} 
        &T.E.& 3 & 14 & 20 & 78 & 380 & 433 & 532 \\
        &Top-1 (\%) & 81.8 &81.8 &81.8 &81.8 &81.8 &81.5 &81.0\\
    \bottomrule
    \end{tabular}
    }
\vspace{-2mm}
    \label{tab:first_n}
\end{table}
\noindent \textbf{Removing first $N$ attention layers.} 
In the main text, we compare NOSE to the random selection strategy. Here, we implement First-$N$ as another baseline, where the first $N$ consecutive attention layers are removed. As shown in~\cref{tab:first_n}, First-$N$ deteriorates quickly with the increase of $N$, while NOSE maintains good performance yet with less transfer entropy (T.E.).

\begin{table}[!t]
    \centering
    \caption{Experiments of removal ratio.}
    \vspace{-0mm}
\resizebox{1.0\columnwidth}{!}
{\tablestyle{3pt}{1.0}
    \begin{tabular}{@{}c|ccccccccccc@{}}
    % \multicolumn{7}{c}{Amount of removed attention layers.} \\
    \toprule
            Num.    &1 &2& 3& 4 &5 &6 &7 &8 &9 &10\\ \hline
            Top-1(\%)   &81.8 &81.8 &81.8 &81.8 &81.8 &81.5 &81.0 &79.4 &76.3 &72.8 \\
    \bottomrule
    \end{tabular}
    }
\vspace{-0mm}
\label{tab:ratio}
\end{table}
\noindent \textbf{Removal rates.} We investigate the removal rates on DeiT-B as in~\cref{tab:ratio}. When it comes to 75\% removal rate (\ie 9 layer), the performance starts to drop drastically.


\end{document}